\documentclass[10pt,twocolumn,letterpaper]{article}

\usepackage[pagenumbers]{cvpr} 
\usepackage{makecell}
\usepackage{algorithm2e}
\newcommand{\methodabbr}{RAIN}
\newcommand{\x}{\textbf{x}}
\newcommand{\y}{\textbf{y}}
\newcommand{\z}{\textbf{z}}

\usepackage[bigfiles]{pdfbase}
\ExplSyntaxOn
\NewDocumentCommand\embedvideos{smm}{
  \group_begin:
  \leavevmode
  \tl_if_exist:cTF{file_\file_mdfive_hash:n{#3}}{
    \tl_set_eq:Nc\video{file_\file_mdfive_hash:n{#3}}
  }{
    \IfFileExists{#3}{}{\GenericError{}{File~`#3'~not~found}{}{}}
    \pbs_pdfobj:nnn{}{fstream}{{}{#3}}
    \pbs_pdfobj:nnn{}{dict}{
      /Type/Filespec/F~(#3)/UF~(#3)
      /EF~<</F~\pbs_pdflastobj:>>
    }
    \tl_set:Nx\video{\pbs_pdflastobj:}
    \tl_gset_eq:cN{file_\file_mdfive_hash:n{#3}}\video
  }
  \pbs_pdfobj:nnn{}{dict}{
    /Type/RichMediaInstance/Subtype/Video
    /Asset~\video
    /Params~<</FlashVars (
      source=#3&
      skin=SkinOverAllNoFullNoCaption.swf&
      skinAutoHide=true&
      skinBackgroundColor=0x5F5F5F&
      skinBackgroundAlpha=0
      autoRewind=true
    )>>
  }
  \pbs_pdfobj:nnn{}{dict}{
    /Type/RichMediaConfiguration/Subtype/Video
    /Instances~[\pbs_pdflastobj:]
  }
  \pbs_pdfobj:nnn{}{dict}{
    /Type/RichMediaContent
    /Assets~<<
      /Names~[(#3)~\video]
    >>
    /Configurations~[\pbs_pdflastobj:]
  }
  \tl_set:Nx\rmcontent{\pbs_pdflastobj:}
  \pbs_pdfobj:nnn{}{dict}{
    /Activation~<<
      /Condition/\IfBooleanTF{#1}{PV}{XA}
      /Presentation~<</Style/Embedded>>
    >>
    /Deactivation~<</Condition/PI>>
  }
  \hbox_set:Nn\l_tmpa_box{#2}
  \tl_set:Nx\l_box_wd_tl{\dim_use:N\box_wd:N\l_tmpa_box}
  \tl_set:Nx\l_box_ht_tl{\dim_use:N\box_ht:N\l_tmpa_box}
  \tl_set:Nx\l_box_dp_tl{\dim_use:N\box_dp:N\l_tmpa_box}
  \pbs_pdfxform:nnnnn{1}{1}{}{}{\l_tmpa_box}
  \pbs_pdfannot:nnnn{\l_box_wd_tl}{\l_box_ht_tl}{\l_box_dp_tl}{
    /Subtype/RichMedia
    /BS~<</W~0/S/S>>
    /Contents~(embedded~video~file:#3)
    /NM~(rma:#3)
    /AP~<</N~\pbs_pdflastxform:>>
    /RichMediaSettings~\pbs_pdflastobj:
    /RichMediaContent~\rmcontent
  }
  \phantom{#2}
  \group_end:
}
\ExplSyntaxOff

\newcommand{\embedvideo}[3]{\embedvideos{\includegraphics[width=#3]{#2}}{#1}}

\definecolor{cvprblue}{rgb}{0.21,0.49,0.74}
\usepackage[pagebackref,breaklinks,colorlinks,allcolors=cvprblue]{hyperref}
\def\paperID{15451} 
\def\confName{CVPR}
\def\confYear{2025}

\title{RAIN: Real-time Animation of Infinite Video Stream}

\author{Zhilei Shu$^{1}$, Ruili Feng$^{2\dag}$, Yang Cao, Zheng-Jun Zha\\\\
\small{$^1$ pscgylotti@gmail.com, $^2$ fengruili.frl@gmail.com, $^\dag$ Project Leader} \\
\url{https://pscgylotti.github.io/pages/RAIN}
}

\begin{document}
\maketitle

\begin{abstract}
Live animation has gained immense popularity for enhancing online engagement, yet achieving high-quality, real-time, and stable animation with diffusion models remains challenging, especially on consumer-grade GPUs.
    Existing methods struggle with generating long, consistent video streams efficiently, often being limited by latency issues and degraded visual quality over extended periods.
    In this paper, we introduce \methodabbr, a pipeline solution capable of animating infinite video streams in real-time with low latency using a single RTX 4090 GPU. 
    The core idea of \methodabbr~is to efficiently compute frame-token attention across different noise levels and long time-intervals while simultaneously denoising a significantly larger number of frame-tokens than previous stream-based methods.
    This design allows \methodabbr~to generate video frames with much shorter latency and faster speed, while maintaining long-range attention over extended video streams, resulting in enhanced continuity and consistency.
    Consequently, a Stable Diffusion model fine-tuned with \methodabbr~in just a few epochs can produce video streams in real-time and low latency without much compromise in quality or consistency, up to infinite long.
    Despite its advanced capabilities, the \methodabbr~only introduces a few additional 1D attention blocks, imposing minimal additional burden.
    Experiments in benchmark datasets and generating super-long videos demonstrating that \methodabbr~can animate characters in real-time with much better quality, accuracy, and consistency than competitors while costing less latency. 
    All code and models will be made publicly available.
\end{abstract}
    
\section{Introduction}
\label{sec:intro}

Live animation has emerged as a powerful tool for enhancing online engagement, bringing characters, avatars, and digital personas to life in real-time. Its growing significance is evident across various domains, from entertainment and gaming to virtual influencers and live-streaming platforms. By enabling dynamic, interactive experiences, live animation fosters more immersive and personalized connections, making it increasingly valuable for social media, online communication, and digital content creation. This demand for engaging real-time animation has sparked interest in developing diffusion models, the most successful image and video generative neural networks, to create smooth, vivid, and responsive animations, especially in applications that require extended live shows or continuous interaction. 

Despite its potential, achieving high-quality, real-time, and stable live animation with diffusion models remains a challenging task, especially when relying on consumer-grade hardware with limited computational power. Current animation methods often require several minutes to generate just a few seconds of video and are incapable of continuously synthesizing long videos that extend for several hours, as commonly needed in practical applications. Consequently, these limitations render most existing animation methods impractical for real-world live animation scenarios.

Recent advances in stream-based diffusion models and diffusion acceleration methods have provided promising pathways toward addressing the challenges of real-time live animation. These methods leverage the multi-step generation nature of diffusion models, allowing stream-based diffusion models to maintain a $\mathrm{StreamBatch}$ of frame tokens corresponding to the number of denoising steps. Each token in this $\mathrm{StreamBatch}$ is incrementally injected with noise levels corresponding to its position, enabling efficient handling of stream inputs and outputting frames in a continuous, streaming fashion. This approach, particularly when combined with acceleration techniques, has significantly improved the speed of frame generation to reach real-time levels.

However, the generation continuity and quality of this process is constrained by the size of the $\mathrm{StreamBatch}$, which is typically limited by the number of denoising steps. Since most acceleration methods sample in fewer than 4 steps, stream-based diffusion models often fail to fully utilize the computational power of even consumer-grade GPUs, thereby limiting their overall performance. Additionally, the relatively small size of the $\mathrm{StreamBatch}$ hinders the model’s ability to compute attention over longer time intervals, which is essential for maintaining continuity in generated video streams. This limitation results in less influence from one frame to the next, reducing the fluidity and consistency of the animation over extended durations. Consequently, existing stream-based methods often struggle with maintaining seamless animation, resulting in latency issues or degraded visual quality, especially during long-duration outputs.

In response to these challenges, this paper introduces \methodabbr, a pipeline solution designed to achieve real-time animation of infinite video streams using consumer-grade GPUs. Unlike previous methods that restrict the $\mathrm{StreamBatch}$ size to match the number of denoising steps, \methodabbr~expands this size by a factor of $p=\frac{\mathrm{GPU\,Capcacity}}{\mathrm{Denoising\,Steps}}$ by assigning every $p$ consecutive frame tokens into denoising groups that share the same noise level, and gradually increasing the noise level across these groups.  This expansion fully utilizes the computational potential of available hardware and enables the model to capture much longer-range temporal dependencies by allowing attention over a larger sequence of frame tokens, significantly improving the consistency and continuity of the generated video streams.
Additionally, while previous methods avoided cross-noise-level attention, we find that it works effectively when combined with the long-range attention across different denoise groups, where each group shares the same noise level. This synergy between long-range attention and cross-noise-level attention significantly boosts continuity and visual quality. By integrating these key elements \methodabbr~achieves substantial improvements in real-time video generation, delivering superior visual quality and consistency over prolonged animations. 

\begin{figure}
    \centering
    \embedvideo{figs/3x3all.mp4}{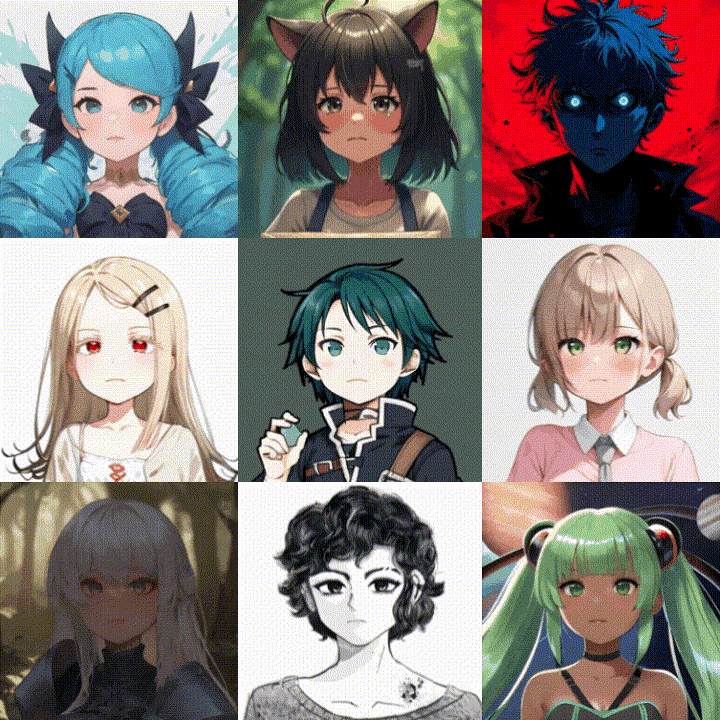}{1\linewidth}
    \caption{Animation clips for crossdomain face morphing. \textit{Best viewed with Acrobat
Reader. Click the images to play the animation clips.}}
    \label{fig:all_teaser}
\end{figure}
\section{Related Work}\label{sec: related_work}
\paragraph{Motion Transfer}
Motion transfer aims at generating image of a character with a set of driven poses. GAN-based methods , \citet{Tulyakov:2018:MoCoGAN, Siarohin_2019_CVPR, Siarohin_2019_NeurIPS, lee2019metapix}, usually tried to inject pose information into intermediate representation of image inside GAN and synthesized edited images. However, due to limitations of GAN itself (e.g. instability and mode collapse), the generated contents suffer from blur, inaccuracy and poor quality. Since the recent success of diffusion model in visual content generation, many works \citep{wang2023disco, xu2023magicanimate, hu2023animateanyone, chang2023magicdance} have delved into the possibility of motion transfer on diffusion model. Since the diffusion model itself has a series of external peripherals, like ControlNet \citep{zhang2023adding}, we can easily make use of different kinds of control signal, like depth, pose, normal and semantic maps. In \citet{wang2023disco}, the input is disentangle into character, motion and background, then the authors make exploits of a ControlNet-like structure and add an additional motion module to keep inter-frame continuity. In \citet{hu2023animateanyone, xu2023magicanimate}, video diffusion model is leveraged for better temporal consistency and two-stage training ensure the disentanglement of motion control, character identity and temporal consistency. These method generally have a long video generation algorithm, that is, performing denoising on overlapped adjacent temporal batches and averaging the overlapped area after every denoising step, or performing generation in a autoregressive manner. However, it requires additional computation for the overlapped frames, while also this is not suitable for the first-in-first-out condition of live stream. 
\paragraph{Stream Video Processing}
For downstream tasks such as live broadcast and online conferences, it is important that the model can process streaming input with infinite length in real-time. StreamDiffusion \citet{kodaira2023streamdiffusionpipelinelevelsolutionrealtime} proposed the ideas of denoising frames with different noisy levels in one batch for better utilization of GPU workload. Live2Diff \citet{xing2024live2diff} used diagonal attention and KV-Cache for real-time video style transfer. StreamV2V 
 \citet{liang2024looking} designed a feature-bank for modeling cross-frame continuity. These works generally adopt the strategy of denoising batches of frames with different noise levels, and make trade off between the quality and efficiency.

\paragraph{Video Style Transfer}
We generally may want to translate images of one specific art style to another. When it comes to videos, continuity and preservation of original objects is usually required. In the early work \citep{gatys2015neuralalgorithmartisticstyle,10.24963/ijcai.2017/310}, Gram matrix is leveraged for minimizing the distance of style between two images. And in \cite{8237429}, normalization layer inside network is adapted for style control. In the recent works with diffusion model, style transfer is achieved generally through IP-Adapter \citep{DBLP:journals/corr/abs-2106-09685} and textual inversion \citep{gal2022textual} together with spatial control signal \citep{zhang2023adding} or latent inversion \citep{song2021denoising}. Recent works \citep{xing2024live2diff, liang2024looking}, use SDEdit \citep{meng2022sdedit} as the image editing backbone and extend the pipeline to match the parallel batch denoising \cite{kodaira2023streamdiffusionpipelinelevelsolutionrealtime} and add additional module for temporal consistency. However, these method heavily depends on temporal module for ensuring the object consistency, while this is inefficient and usually outputs unstable result.

\section{Preliminaries}

\subsection{Consistency Model}
Consistency Distillation \cite{Song2023ConsistencyM} is an efficient way for diffusion model sampling acceleration. A Consistency Model $\boldsymbol f(x,t)$ satisfies:
\begin{equation}
    \boldsymbol f(\textbf{x}_{\epsilon},\epsilon) = \textbf{x}_{\epsilon},
\end{equation}
namely, $\boldsymbol f(\cdot,\epsilon)$ is an identity function for certain $\epsilon\sim 0$ ($\epsilon$ is close enough to $0$, for numerical stability, not necessary $0$).
For a diffusion model with form in SDE:
\begin{equation}
    \text{d}\textbf{x}_t =\boldsymbol{\mu}(\textbf{x}_t,t) \text{d} t+\sigma(t)\text{d}\textbf{w}_t,
\end{equation}
and we have its Probability Flow (PF) ODE\citep{song2021scorebased}:
\begin{equation}
    \text{d}\textbf{x}_t =\left\lbrack\boldsymbol{\mu}(\textbf{x}_t,t) -\frac12\sigma^2(t)\nabla\log p_t(\textbf{x}_t)\right\rbrack\text{d}\textbf{w}_t.
\end{equation}
With an initial noise $\textbf{x}_T\sim\mathcal N(\boldsymbol{O},\boldsymbol{I})$, PF-ODE basically determines a trajectory for $t\in\lbrack0,T\rbrack$, denoted by $(\textbf{x}_t, t)$. We expect that for every trajectories $(\textbf{x}_t, t),t\in\lbrack\epsilon,T\rbrack$
\begin{equation}
    \boldsymbol f_{\boldsymbol{\theta}}(\textbf{x}_t,t)=\textbf{x}_{\epsilon}
\end{equation}
So that the model $\boldsymbol f_{\boldsymbol{\theta}}$ can generate sample in one step. Generally, we have the following consistency distillation loss \cite{Song2023ConsistencyM}:
\begin{align}
    \mathcal L_{\text{CD}}(\boldsymbol{\theta},&\boldsymbol{\theta}^-;\phi):=\nonumber\\\mathbb E&\left\lbrack\lambda(t_n)d\left(\boldsymbol f_{\boldsymbol{\theta}}(\textbf{x}_{t_{n+1}},t_{n+1}),\boldsymbol{f}_{\boldsymbol{\theta}^-}(\hat{\textbf{x}}_{t_n}^{\phi},t_n)\right)\right\rbrack,
\end{align}
where $\phi$ is an ODE solver with original diffusion model, $d$ is an arbitrary distance metric, $\boldsymbol{\theta}^-$ is an exponential moving average (EMA) of $\boldsymbol{\theta}$.\\
The distilled model can then be used for fast sampling. Multi-steps sampling is achieved through iteratively predicting $\textbf{x}_{\epsilon}$ from $\textbf{x}_{t_{n+1}}$ and adding noise to next timesteps to gain $\textbf{x}_{t_{n}}$.  Usually in $4$ steps, the result are good enough.
\subsection{Stream Diffusion}
Due to the nature of multistep sampling of diffusion model, drawing a sample from diffusion model generally requires several steps of function evaluation. Under the scene of streaming video processing, we intrinsically want to make full use of GPU with batch computing for acceleration. StreamDiffusion \cite{kodaira2023streamdiffusionpipelinelevelsolutionrealtime} firstly proposed to push frame with different noise level into one batch. Additionally, StreamDiffusion adopted LCM Acceleration \cite{luo2023latent,luo2023lcmlorauniversalstablediffusionacceleration} and TinyVAE\cite{madebyollintinyvae2023} for further acceleration. We also include these optimizations in our method.
\subsection{Reference Mechanism}
The Reference Mechanism proposed in AnimateAnyone \citep{hu2023animateanyone} tries to preserve character identity for 2D UNet model with a reference image. Initially, a pretrained 2D UNet is leveraged as a ReferenceNet. The ReferenceNet will perform an inference on the reference image and we cache the input hidden states before every spatial attention operation. We can then use these hidden states as reference information. Assuming we are generating images with a denoising 2D UNet which shares the same architecture as ReferenceNet. Then before every spatial self-attention operation in denoising UNet, we concatenate the corresponding reference hidden states with original Key and Value inputs. Formula \ref{formula:reference} displays the detailed operation.

\begin{equation}
    \boldsymbol{Q}=\boldsymbol{W^{Q}}\boldsymbol{X},\boldsymbol{K}=\boldsymbol{W^{K}}\left\lbrack\boldsymbol{X},\boldsymbol{Z}\right\rbrack,\boldsymbol{V}=\boldsymbol{W^{V}}\left\lbrack\boldsymbol{X},\boldsymbol{Z}\right\rbrack,
    \label{formula:reference}
\end{equation}
where $\boldsymbol X$ is the hidden states of denoising UNet, $\boldsymbol Z$ is the corresponding reference hidden states. It is noticed that the reference mechanism generally double the cost of spatial attention operation. Also we can have multiple guidance images at once.

\section{Method}\label{sec: method}
\begin{figure*}[htbp]
\includegraphics[trim=0 70 0 70,clip,width=1.0\linewidth]{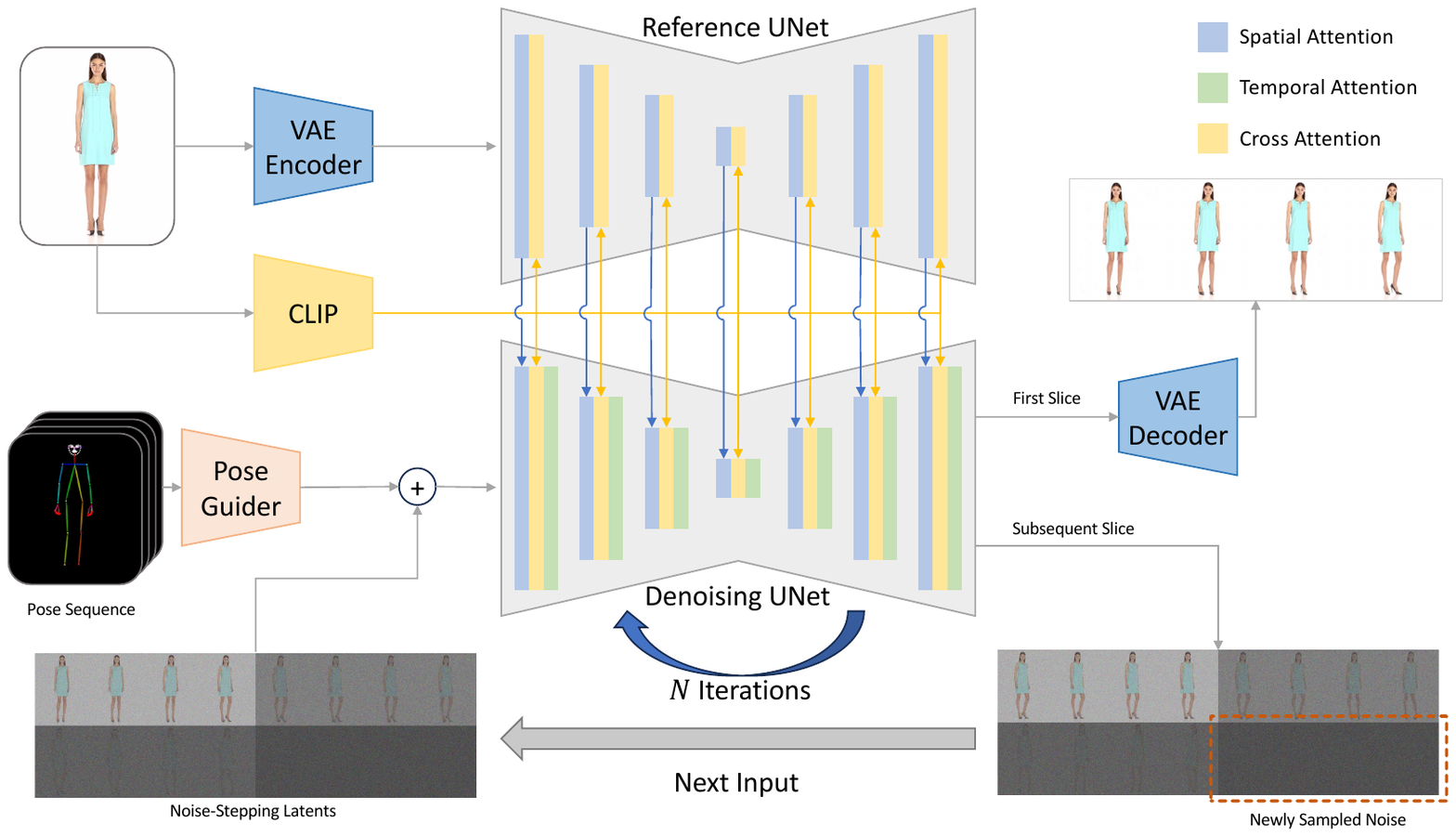}
\centering
\caption{The overview pipeline of \methodabbr. We first feed the reference image into Reference UNet and CLIP Text Encoder, the spatial attention feature from Reference UNet and CLIP embeddings are fed into Denoising UNet. The pose sequence is mapped through pose guider and added to the intermediate feature after post convolutional layer of Denoising UNet. Every times after $N$ iterations of UNet function calls, the noise level of each frames is reduced by $T/p$ steps, and the first $K/p$ frames are already clean. We \textbf{pop} out first $K/p$ frames and \textbf{push} $K/p$ frames of standard noise to the latent piles. Each clean latent is then decoded by VAE Decoder as a video frame.}\label{fig:pipeline}
\end{figure*}

In the task of Human Image Animation, we want to generate videos of a given character (in images) according to a pose sequence. There are two key requirements for the visual quality: \textbf{character consistency} and \textbf{temporal continuity}. The existing framework, based on diffusion model, solves the consistency problem through Reference mechanism, and the temporal continuity is ensured by an additional 1D temporal attention module. However, for making much practical use, we now consider the input as a streaming video, and there are additional requirements for \textbf{latency} and \textbf{fps}. Here we are going to give a detailed explanation of our \methodabbr~ pipelines, the overall framework of \methodabbr~is presented in Figure \ref{fig:pipeline}. Our pipeline is mainly adapted from AnimateAnyone \cite{hu2023animateanyone}.

\subsection{Temporal Adaptive Attention}

For better processing of the streaming video input, we make some changes on the temporal attention parts of a given 2D + 1D Diffusion Model. Generally, for every UNet inference step, assuming there are $K$ frames in one batch, we separate them evenly into $p$ group (satisfying $p\mid K$). The noise level for frames in each group increases in a step-by-step manner like the StreamDiffusion. More specifically, for Motion Module in AnimateDiff \citep{guo2023animatediff}, we have $K=16$. And in order to be compatible with the $4$ - step sampling of LCM, we choose $\frac Kp=4$, namely $p=4$. We fix $T=1000$ as it is in the Stable Diffusion, the noise level (represented in timesteps) of each frames shall be:
\begin{align}
    \begin{matrix}
        \lbrack &t_0, &t_0, & t_0, & t_0, \\
                &t_0 + 250, &t_0 + 250, &t_0 + 250, &t_0 + 250, \\
                &t_0 + 500, &t_0 + 500, &t_0 + 500, &t_0 + 500, \\
                &t_0 + 750, &t_0 + 750, &t_0 + 750, &t_0 + 750&\rbrack
    \end{matrix}~,\label{formula:timesteps}\\~~t_0\in\lbrack1,250\rbrack.\nonumber
\end{align}
Which means the noise level difference between every adjacent frame groups is $250$ steps.

\subsection{Train and Inference}

The Training of \methodabbr~ adopts the two-stage strategy in \citet{hu2023animateanyone}. In the first stage, models are trained on image pairs from same videos, the reference net and pose guider are trained in this stage together with the denoising unet.
For the second stage, we sample $K$ frames of video and add noise according to the timesteps in \ref{formula:timesteps}. Here we chosing $t_0$ ranging evenly from $1$ to $\frac Tp$. 

We only finetune the motion module on these frames with non-uniform noise level , and we call this procedure as forcing the motion module to be \textbf{temporal adaptive}. Then the denoising model can accept stream video inputs and process infinite long videos.

For sampling with $p\cdot N$ steps, $t_0$ iteratively takes value of $\frac Tp, \frac{(N-1)T}{Np},\cdots,\frac{2T}{Np},\frac{T}{Np}$, every denoising step will remove $\frac{T}{Np}$ steps of noise from each frames. If first $p$ frames is already clean, we remove them and append $p$ frames of standard noise to the bottom. For $4$ - step LCM sampling, $N$ actually equals $1$.

Initially, we only inference with $p$ frames of pure noise, and after $N$ steps, we push another $p$ frames of pure noise at bottom. We repeat this procedure until all frames is fulfilled. This soft startup strategy can be of some instability, but soon it will reach stable. We notice that simply filling the buffer with one still image for startup usually outputs degenerate results and has more error accumulation at start.

\subsection{LCM Distillation}

In order to achieve real-time inference, we adopt Consistency Distillation \cite{Song2023ConsistencyM, luo2023latent}, which can speed up inference by 5x - 10x compared with DDIM Sampling. We adopt the 3D Inflation Initialization Strategy proposed in the AnimateLCM \cite{wang2024animatelcm}, which first performs consistency distillation on 2D UNet. Then, for 3D consistency distillation, the 3D online student model is initialized with 2D LCM and Motion Module, while the 3D target model is initialized with original 2D UNet and Motion Module. We also absorbs the classifier-free guidance \cite{ho2022classifierfreediffusionguidance} functionality into the distilled model for further acceleration. For different datasets, the best guidance strength $\omega$ ranging from $2.0$ to $3.5$.

\subsection{Architecture}
We choose a variant of Stable Diffusion \citep{Rombach2021HighResolutionIS}, namely the SD-Image Variations \citep{sdimagevariation}, as the base model. This variant is finetuned on CLIP \cite{Radford2021LearningTV} prompts and adopts $v$-prediction \cite{DBLP:journals/corr/abs-2202-00512}. The reference net shares the same structure as the base model, while the last Up Block of UNet is removed since it is never used. We choose the AnimateDiff Motion Module \cite{guo2023animatediff} as 1D temporal block. The pose guider is a simple convolutional network that map the input with shape $3\times H\times W$ to the feature shape $320\times\frac{H}8\times\frac{W}8$.

\section{Experiments}\label{sec: exps}
We perform \methodabbr~ on several downstream tasks under the video live stream scene. Different control signals are used for video generation. Models are trained on 8x Nvidia A100 GPU. For the first image training stage, batch size is set to $32$. And during the second video training stage, batch size is set to $8$. Videos are sampled into clip with length of $16$, and one frame from same video is randomly selected as reference image. The temporal adaptive arguments $p$, $K$ are specified to $4$. For consistency distillation stage, we use Huber loss with $c =0.001$ and DDIM Solver with $100$ timesteps, while batch size is set to $8$ for both stage. We use AdamW \cite{loshchilov2018decoupled} with learning rate of $10^{-5}$ in all training stage. It is worth noting that in the inference stage, we achieve $18$ fps on single RTX 4090 GPU for $512\times512$ videos with TensorRT Acceleration and TinyVAE (including the processing time of DWPose). 
\begin{figure*}[htbp]
\includegraphics[width=1.0\linewidth]{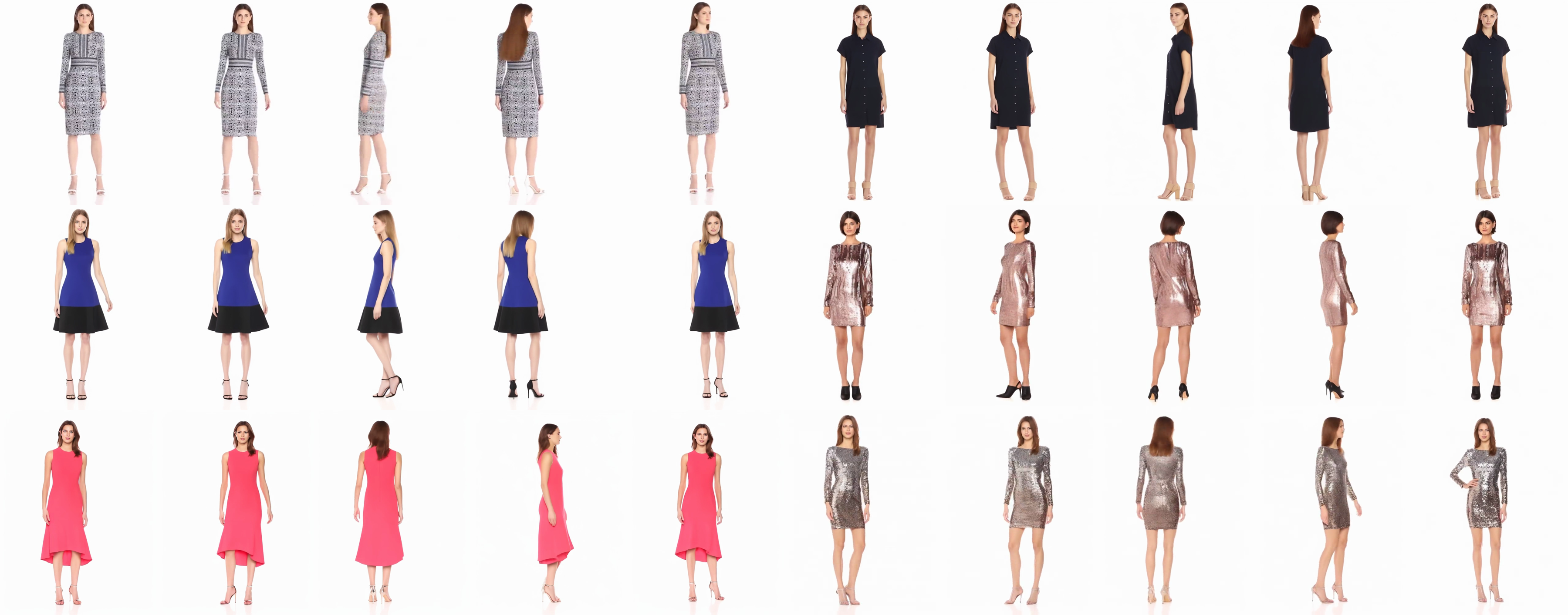}
\centering
 \caption{Generation results from UBC-Fashion test dataset.}\label{fig:fashion}
\end{figure*}
\subsection{Human Whole body Movement Generation}
\paragraph{Dataset}
We use UBC-Fashion Dataset \cite{DBLP:journals/corr/abs-1910-09139} which consists of $500$ training videos and $100$ testing videos with clean background. Different characters in the videos wear different clothes for display. 
\paragraph{Settings}
We choose DWPose \cite{yang2023effective} as whole body keypoints extractor. Video frames are resize to $512\times 768$. We train the model with $30$k steps for first stage and $20$k steps for second stage. For consistency distillation, both 2D and 3D model are trained with $1200$ step. The guidance strength is set to $2.0$.
\paragraph{Results}
We compare our result with previous works to see whether the generation quality harms significantly as a price for the acceleration. The quantitative results is displayed in the Table \ref{tab:fashion}, and some generation results from the test set is displayed in Figure \ref{fig:fashion}. We take PSNR \cite{5596999}, SSIM \cite{1284395}, LPIPS \cite{8578166} and FVD \cite{unterthiner2019accurategenerativemodelsvideo} as metrics. Evaluation results indicates that the quality of \methodabbr~ generation does not decrease too much (compared with AnimateAnyone).
\begin{table}[h]
\small
\centering
\begin{tabular}{lcccc}
\hline
Method        & PSNR $\uparrow$ & SSIM $\uparrow$ & LPIPS $\downarrow$ & FVD $\downarrow$ \\ \hline
MRAA\cite{siarohin2021motion}          & -                                   & 0.749                               & 0.212                                  & 253.6                                \\
TPSMM\cite{Zhao2022ThinPlateSM}         & -                                   & 0.746                               & 0.213                                  & 247.5                                \\
BDMM\cite{10378041}          & 24.07                               & 0.918                               & 0.048                                  & 168.3                                \\
DreamPose\cite{karras2023dreamposefashionimagetovideosynthesis}     & -                                   & 0.885                               & 0.068                                  & 238.7                                \\
DreamPose*        & 34.75*                              & 0.879                               & 0.111                                  & 279.6                                \\
AnimateAnyone & \textbf{38.49*}                              & \textbf{0.931}                               & \textbf{0.044}                                  & \textbf{81.6}                                 \\
\methodabbr   & 23.99                               & 0.921                               & 0.063                                  & 85.2                                 \\ \hline
\end{tabular}
\caption[Table]{Quantitative results on UBC-Fashion Dataset. DreamPose* indicates results without finetuning. These PSNR values above $30$ are the results of overflow. Their authors use code that treat image array as uint8 type\footnotemark. If the overflowed algorithm is used, the PSNR value of \methodabbr~is 37.20.}
\label{tab:fashion}
\end{table}
\footnotetext{https://github.com/Wangt-CN/DisCo/issues/86}
\subsection{Cross Domain Face Morphing}
\begin{figure*}[htbp]
\includegraphics[width=1.0\linewidth]{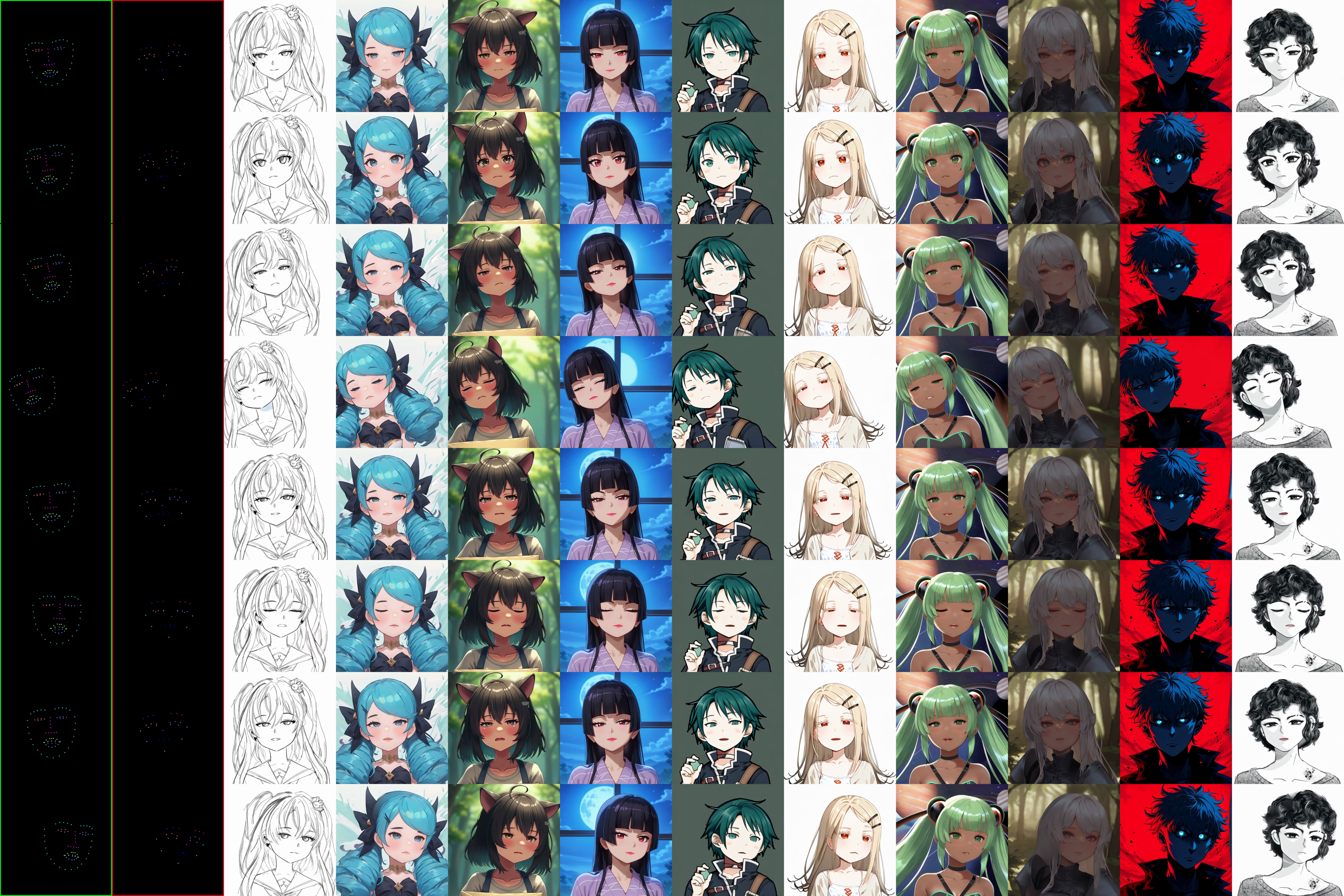}
\centering
 \caption{Results of cross domain face morphing: the two leftmost columns are the original DWPose sequence and the transformed landmarks. Characters' expressions follows the input exactly. However, for different characters and humans, the transformation parameters need to be adjusted accordingly. For example, the length of face and size of eyes are varying for different characters. }\label{fig:animeface}
\end{figure*}

\paragraph{Dataset}
We collect $1.8$k video clips of anime face from YouTube as training datasets. Their length ranging from $3$ seconds to $20$ seconds and their aspect ratio is cropped to approximately $1.0$.
\paragraph{Settings}
We use DWPose \cite{yang2023effective} as facial landmarks extractor for inference (on real human face), and use AnimeFaceDetector \cite{anime-face-detector} to annotate the dataset. In order to address the gap between these two kinds of annotations, we design a composition of simple linear transformations. Through which landmarks of real human (DWPose) are mapped to landmarks of anime face (AnimeFaceDetector). After the transformations, the opening and closing of the eyes and mouth can still be maintained, so it can be used to control the generation. In our test, DWPose can directly locate anime face with a certain precision, but there are usually gaps in eyes and mouth. Model directly trained on DWPose annotated datasets is insensitive to eye and mouth movements. Video frames are resize to $512\times 512$. We train the model with $60$k steps for first stage and $30$k steps for second stage. For consistency distillation, both 2D and 3D model are trained with $1200$ step. The guidance strength is set to $2.5$.
\paragraph{Results}
We show some cases for cross domain face morphing in the Figure \ref{fig:animeface}. Head position and expressions are successfully ported to the anime character.
\subsection{Style Transfer}
\begin{figure*}[htbp]
\includegraphics[width=1.0\linewidth]{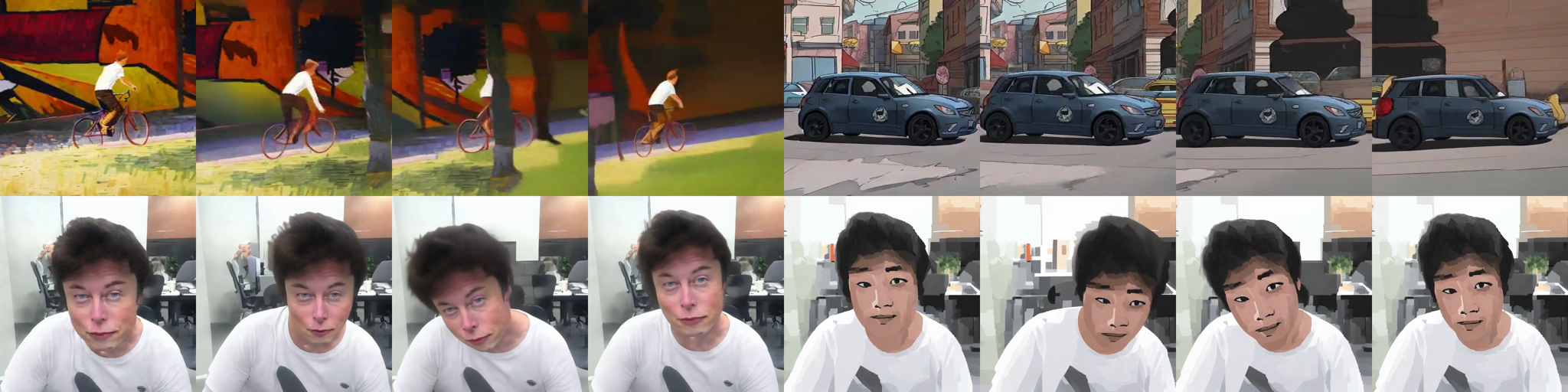}
\centering
 \caption{Results of style transfer: Dynamic scenes lead to gradual loss of detail and synthesis failure, while stable scenes can be synthesized normally.}\label{fig:style}
\end{figure*}

\paragraph{Dataset}
We use a $50$k randomly selected video clips subset of Panda-70M Dataset \cite{chen2024panda70m} as the training dataset. Panda-70M consists of high quality video clips of various scenes and topics. 
\paragraph{Settings}
We choose MiDaS \cite{Ranftl2021,Ranftl2022} as the depth estimator. We randomly crop a patch from the video clips and resize it to $512\times 512$. We train the model with $60$k steps for first stage and $40$k steps for second stage. For consistency distillation, both 2D and 3D model are trained with $1200$ step. The guidance strength is set to $3.5$. For inference, the original image will firstly be transferred to target style through SDEdit \cite{meng2022sdedit} and ControlNet \cite{zhang2023adding}, and then used as reference image.
\paragraph{Results}
We show some cases for video style transfer in the Figure \ref{fig:style}. Previous works mainly compare several metrics on DAVIS-2017 \cite{Pont-Tuset_arXiv_2017} which consists of videos with dynamic background. Since our method does not apply to scene with highly motion, we only compare FPS with these works which focuses on live stream processing in Table \ref{tab:styletransfer}.
\begin{table}[h]
\small
\centering
\begin{tabular}{lccccc}
\hline
Method & FPS $\uparrow$   \\ \hline
StreamDiffusion & \textbf{37.13}                                 \\
Live2Diff                                   & 16.43                                 \\
\methodabbr                                & 18.11                                \\\hline
\end{tabular}
\caption{FPS comparison for live stream tasks. (Single RTX 4090)}
\label{tab:styletransfer}
\end{table}
\begin{figure*}[htbp]
\includegraphics[width=1.0\linewidth]{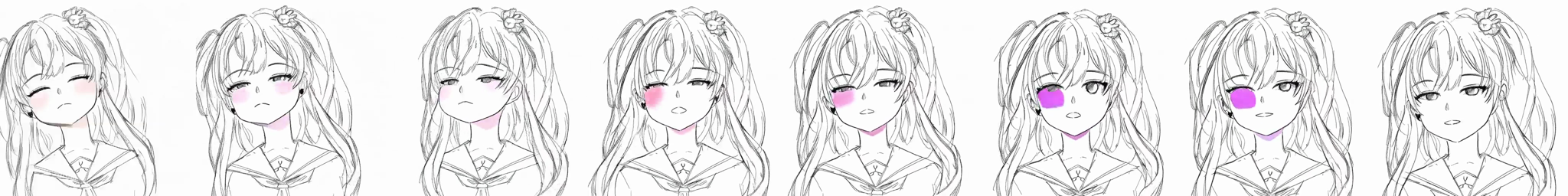}
\centering
 \caption{An error accumulation case: The abnormal blush that initially occurs causes exceptional color blocks in the following generation results. Finally it disappears after $600$ frames.}\label{fig:error_accumulation}
\end{figure*}
\subsection{Ablation Studies}
\paragraph{Temporal Batch Size} By default the temporal batch size $K$ is set to $16$ and we use $4$-step LCM sampling. Every $\frac{K}p=4$ frames forms a group that shares the same noise level. Here we try to reduce $\frac{K}p$ to $2$, namely $K=8$, and examine the result. Table \ref{tab:temporalwindowsize} shows the quantitative results with different $K$. It can be seen that PSNR, SSIM, LPIPS do not harms significantly while FVD gets a lot worse. Lack of previous frames that could be seen results in bad temporal continuity.
\begin{table}[h]
\small
\centering
\begin{tabular}{lccccc}
\hline
$K$        & Masked & PSNR $\uparrow$ & SSIM $\uparrow$ & LPIPS $\downarrow$ & FVD $\downarrow$ \\ \hline
$16$ & False& \textbf{23.99}                              & \textbf{0.921}                               & \textbf{0.063}                                  & \textbf{85.2}                                 \\
$8$   & False & 23.72                               & 0.919                               & 0.064                                  & 145.8                                 \\
$16$   & True & 23.63                               & 0.918                               & 0.066                                  & 284.5                                \\\hline
\end{tabular}
\caption{Quantitative results on UBC-Fashion Dataset with different temporal batch sizes and mask strategies. `True' and `False' of `Masked' denote whether a causal mask is applied to temporal attention module.}
\label{tab:temporalwindowsize}
\end{table}
\paragraph{Attention Mask} In \methodabbr~pipeline, the temporal attention is taken over the entire temporal batch. While generally in models like LLM, auto-regressively processing of a sequence usually requires a causal mask. Intuitively, subsequent frames should not have effect on previous frame. Here we train and inference on a model that accepts causal mask input. The results are shown in Table \ref{tab:temporalwindowsize}. The temporal continuity gets extremely worse, every $4$ frame there will be an obvious jitter. such phenomena is not observed in model before distillation (which mean that videos sampled with $20$ step DDIM do not exhibit obviously jittering).
\section{Discussions and Conclusions}
\subsection{Limitations}
\paragraph{Error Accumulation} Rarely, a little stir in previous frame may cause the subsequent frames output images with abnormal color. Character needs to appear in and out of the camera so that the synthesized results can return to normal. The streaming structure of \methodabbr~ causes the error to accumulated with time. Figure \ref{fig:error_accumulation} shows a case for a small error accumulating into a weird generation result. We will try to fix this by reducing the influence strength of temporal attention.
\paragraph{Fixed Scene} Since most of the details of the main object are provided by the reference net, the occluded areas and new objects may have a noisy texture and exhibit degenerate results. One can fix this by add more different reference images, but this reduce the fps. However, for tasks like live broadcasts and online meetings, the scene is usually fixed. So there is no problem with these tasks.
\subsection{Potential Influence}
Although we will not provide a version on synthesizing video on real human faces domain, the method still can be used to generate fake face videos of real human. However, some fake image detection method \cite{sha2023defakedetectionattributionfake, belli2022onlineadaptivepersonalizationface, wang2023dire} can be used to identify these generated results.
\subsection{Conclusion}
In this paper, we propose \methodabbr, a pipeline for real-time infinite long video stream animation. Compared with previous works, we relax the excessive dependence on temporal attention and use spatial attention which can provide more stable details. We get more consistent, stable and smooth result compared with previous works. We perform experiments on several attractive tasks. Enabling the practical applications for downstream tasks like live broadcast and online conferences. There are also more applications for online spiritual entertainment like virtual youtuber and online virtual chat. We allow the users to transform into their beloved virtual characters in real-time. And we will try to implement this in a more interactive way.
\section{Acknowledgment}
Our work is based on AnimateAnyone\cite{hu2023animateanyone}, and we use the code from Moore-AnimateAnyone\cite{mooreanimateanyone}, Open-AnimateAnyone\cite{guoqincodeopenanimateanyone}, TinyAutoencoder\cite{madebyollintinyvae2023}\cite{madebyollintaesdv} and AnimeFaceDetector\cite{anime-face-detector}, DWPose\cite{yang2023effective}. Thanks to these teams/authors for their work.

Special thanks to CivitAI Community\footnote{https://civit.ai} and YODOYA\footnote{https://www.pixiv.net/users/101922785} for example images. Thanks to Jianwen Meng\footnote{jwmeng@mail.ustc.edu.cn} for pipeline design.
{
   \small
   \bibliographystyle{ieeenat_fullname}
   \bibliography{main}
}

\clearpage
\setcounter{page}{1}
\maketitlesupplementary

\section{Implementation Details}
\begin{figure}[htbp]
\includegraphics[width=1.0\linewidth]{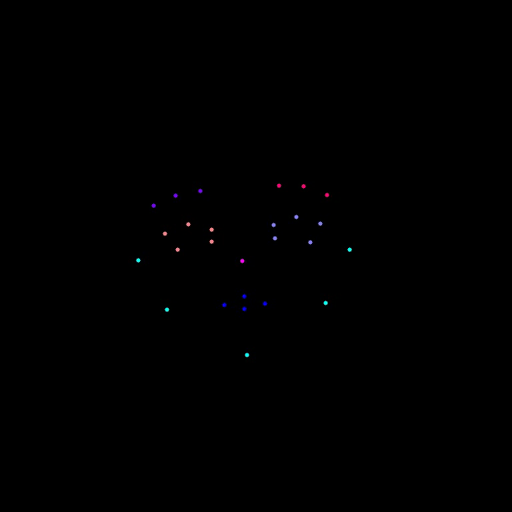}
\centering
 \caption{Anime Facial Landmarks with 26 points}\label{fig:animelandmarks}
\end{figure}

\begin{figure*}[htbp]
\includegraphics[width=1.0\linewidth]{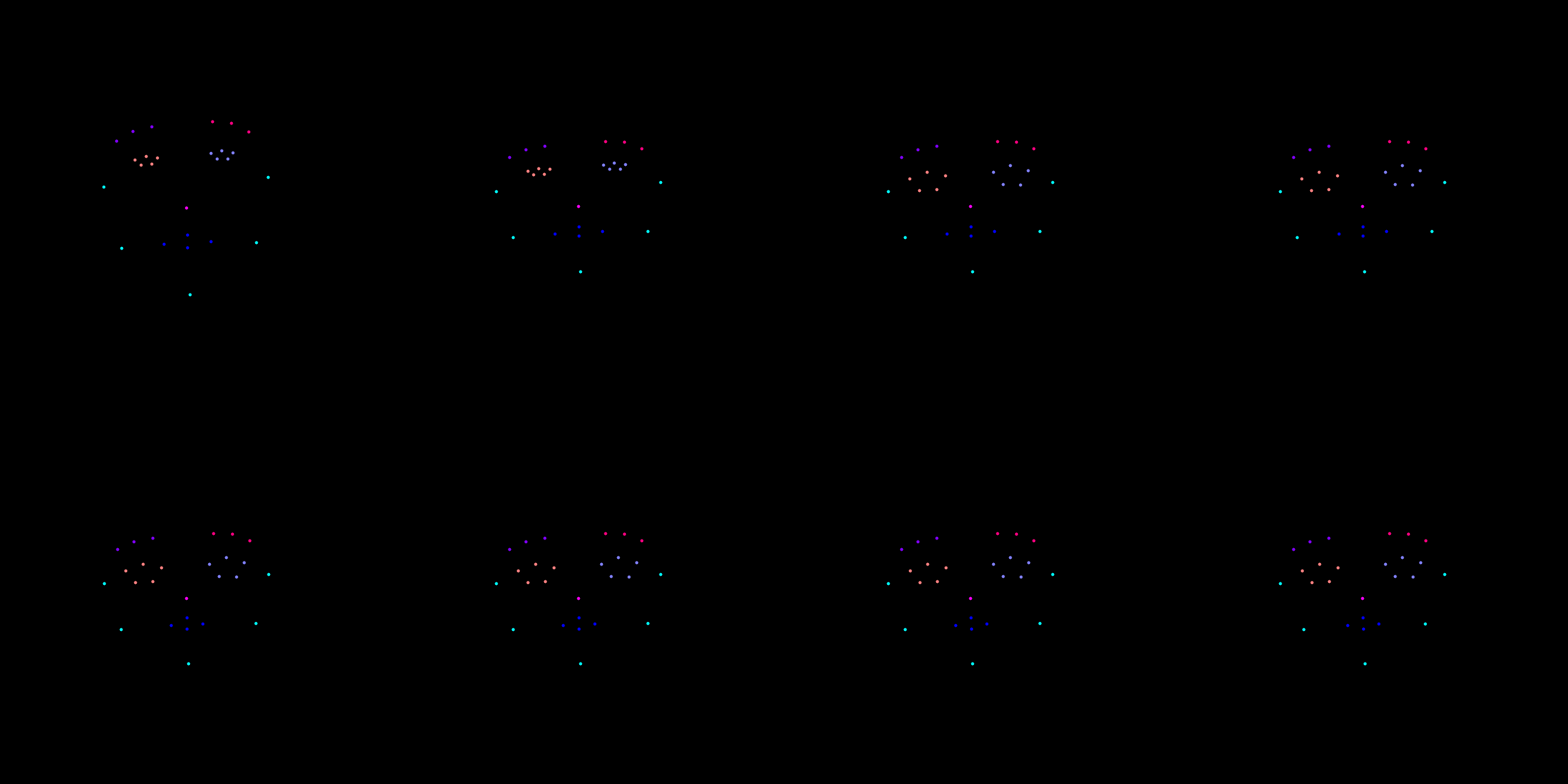}
\centering
 \caption{Transformations of Landmarks: from left top to right bottom, landmarks from humans are mapped to landmarks of anime characters.}\label{fig:mapping}
\end{figure*}

\paragraph{Inference}
We show the inference pipeline of our \methodabbr~in Algorithm \ref{algo:infer}. The basic setting is $K=16,p=4,H=W=512,N=1$. The algorithm can inference with streaming video input with infinite length. LCM scheduler is used for sampling. The temporal batch is not full in first and last $\frac{K}p-1$ iterations. This soft start strategy can benefit the stability at the beginning.
\begin{algorithm*}[ht]
 \SetKwData{Left}{left}\SetKwData{Right}{right}\SetKwData{This}{this}\SetKwData{Up}{up}
 \SetKwFunction{Union}{Union}\SetKwFunction{FindCompress}{FindCompress}
 \SetKwInOut{Input}{input}\SetKwInOut{Output}{output}
 
 \Input{Temporal batch size $K$, temporal group size $p$, denoising step per group $N$, video with length $L$ (satisfying $p \mid L$): $\x_0\in \text{ShapeLike}(L,W,H,C)$, reference image $\y\in\text{ShapeLike}(H,W,C)$, sampling scheduler $s$, pose extractor $e$, pose guider $g$, reference UNet $U_r$, denoising UNet $U_s$, VAE Encoder and Decoder $V_e, V_d$}
 \Output{Processed video $\z_0\in\text{ShapeLike}(L,W,H,C)$}
 \BlankLine
  Intermediate Feature $\textbf{c}\gets g(e(x_0))\in\text{ShapeLike}(L,\frac W8,\frac H8,C')$\\
  Noisy Latents $\z\sim\mathcal N(\textbf{O},\textbf{I})\in\text{ShapeLike}(L,\frac W8,\frac H8,C')$\\
  Reference Attention Features $\textbf{f}\gets U_r(V_e(\y))\in\text{List}\left\lbrack\text{ShapeLike}(L_i,C_i)\right\rbrack$\\
  Total Group Amount $w\gets\frac{T}{p}$\\
  Temporal Adaptive Steps $a \gets\frac{K}p$\\
  Step Length $l\gets \frac{T}{Na}$\\
 \For{$i\leftarrow 1$ \KwTo $w + a -1$}{
  \For{$j\leftarrow 1$ \KwTo $N$}{\label{forins}
   \Left$\gets\max(0, i - a) * p$\\
   \Right$\gets \min(w, i) * p$\\
   Timestep $t\gets\left\lbrack\underbrace{\frac{T}a,\cdots,\frac Ta}_{\times p},\underbrace{\frac{2T}{a},\cdots,\frac{2T}{a}}_{\times p},\cdots,\underbrace{T,\cdots,T}_{\times p}\right\rbrack-l\times(j-1)$\\
   $\boldsymbol\varepsilon\gets U_s(\z\lbrack\Left:\Right\rbrack,\textbf{c}\lbrack\Left:\Right\rbrack,\textbf{f},t) $\\
   $\z\lbrack\Left:\Right\rbrack\gets s(\z\lbrack\Left:\Right\rbrack,\boldsymbol{\varepsilon},t)$
   
  }
 }
 \textbf{return} $V_d(\z)$
 \caption{streaming video processing}\label{algo:infer}
\end{algorithm*}
\paragraph{Dataset} The Anime Face Dataset is privately collected on YouTube. We download anime style videos and crop them manually into 1.8k clips with aspect ratio of approximately $1.0$. We intentionally select clips with simple and static background, various styles and directly facing. We use the AnimeFaceDetector to annotate the dataset with $26$ facial landmarks (Figure \ref{fig:animelandmarks}).
\begin{figure}[htbp]
\includegraphics[width=1.0\linewidth]{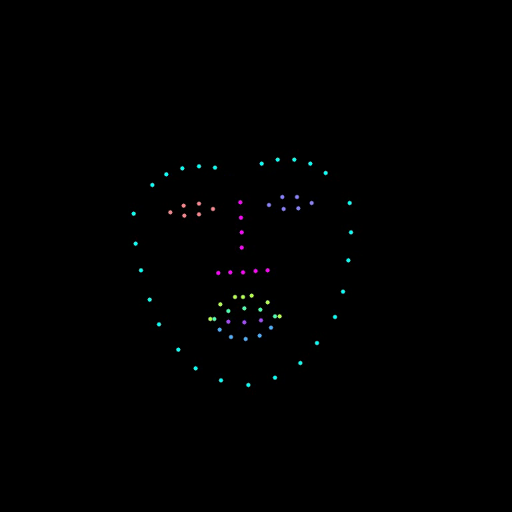}
\centering
 \caption{Real Human Facial Landmarks with 68 points}\label{fig:reallandmarks}
\end{figure}
\begin{figure}[htbp]
\includegraphics[width=1.0\linewidth]{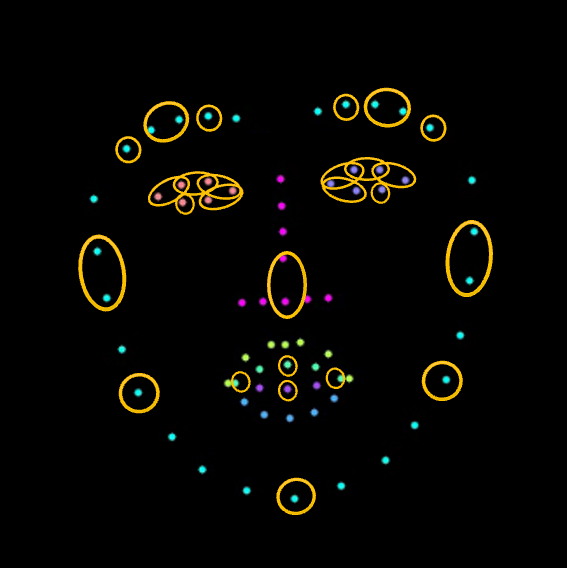}
\centering
 \caption{Selected and Merged Facial Landmarks}\label{fig:merged}
\end{figure}
\paragraph{Keypoint Transformations} In order to match the result of DWPose (Figure \ref{fig:reallandmarks}) to AnimeFaceDetector, we first select specific landmarks from DWPose (Figure \ref{fig:merged}). In the figure, every yellow circle represents a point group, and we use the average of the groups as mapped points. Then we apply linear transformations to keep the results consistent with the AnimeFaceDetector (Figure \ref{fig:mapping}).

\end{document}